\documentclass[conference]{IEEEtran}
    \IEEEoverridecommandlockouts
    
    \usepackage{cite}
    \usepackage{amsmath,amssymb,amsfonts}
    \usepackage{algorithmic}
    \usepackage{graphicx}
    \usepackage{textcomp}
    \usepackage{booktabs}
    \usepackage{xcolor}
    \usepackage{hyperref}
    \usepackage{subcaption}
    \usepackage{float}
    \usepackage{soul}
    \def\BibTeX{{\rm B\kern-.05em{\sc i\kern-.025em b}\kern-.08em
    		T\kern-.1667em\lower.7ex\hbox{E}\kern-.125emX}}
    
    \hypersetup{
    	colorlinks=true,
    	citecolor=blue,
    	linkcolor=blue,
    	urlcolor=blue
    }
    
\begin{document}
    	
    	\title{Learning-Based Navigation for Indoor Mobile Robots}
    	
    	\author{
    	\IEEEauthorblockN{
    		Tri-Tin Nguyen,
    		Tien-Dat Nguyen,
    		Gia-Uy Le,
    		Vinh Nguyen,
    		Vinh-Hao Nguyen\IEEEauthorrefmark{1}
    	}
    	\IEEEauthorblockA{
    		Faculty of Electrical and Electronic Engineering, Ho Chi Minh City University of Technology, VNU-HCM \\ Ho Chi Minh City, Vietnam
    	}
    	\IEEEauthorblockA{
    		tin.nguyen2004@hcmut.edu.vn,  nguyentiendat.sdh242@hcmut.edu.vn,
            uy.legia7@hcmut.edu.vn, \\ vinh.nguyen2604@hcmut.edu.vn, 
    		vinhhao@hcmut.edu.vn 
    	}
    	\IEEEauthorblockA{\IEEEauthorrefmark{1}Corresponding author.}
    	}
    
    	\maketitle
    	
    	\begin{abstract}
    		This paper presents a learning-based navigation framework for indoor mobile robots. The proposed method combines a supervised neural global planner, trained from cost-aware A* expert trajectories, with the proposed Learning-Based DWA local planner, which is formulated as discrete candidate selection over the Dynamic Window Approach (DWA) action lattice. For local planning, the policy is first trained by behavior cloning and then refined by Proximal Policy Optimization (PPO) under feasibility-aware masking. The framework is implemented and evaluated in both simulated and real-world indoor environments. Experimental results show that the proposed method generates feasible global routes and reliable local motion commands for safe goal-directed navigation in the presence of obstacles. These results demonstrate the effectiveness of integrating learning-based global planning with reinforcement-learning-refined local control for indoor mobile robot navigation. The source code will be released at \url{https://ntdathp.github.io/rl_robot_web/}.
    	\end{abstract}
        
        
\section{Introduction}
Indoor mobile robot navigation commonly follows a hierarchical architecture, in which a global planner generates a map-level route and a local planner produces real-time commands for obstacle avoidance. However, classical planners rely on handcrafted search rules and may produce conservative motions in cluttered environments.
        
To address these limitations, this paper proposes a learning-based navigation framework that combines a supervised neural global planner trained from cost-aware A* expert trajectories with a learning-based local planner formulated as discrete candidate selection over the DWA action lattice. The local policy is first trained by behavior cloning and then lightly refined by PPO under feasibility-aware masking, yielding a unified framework for map-level route generation and structured local motion control.

This work contributes a deployable hybrid learning-based navigation framework that preserves DWA feasibility by selecting actions from the DWA lattice rather than directly learning unconstrained continuous velocity commands.

\begin{figure}[htbp]
    \centering
    \includegraphics[width=1.15\columnwidth, trim=2.5cm 0 0 0, clip]{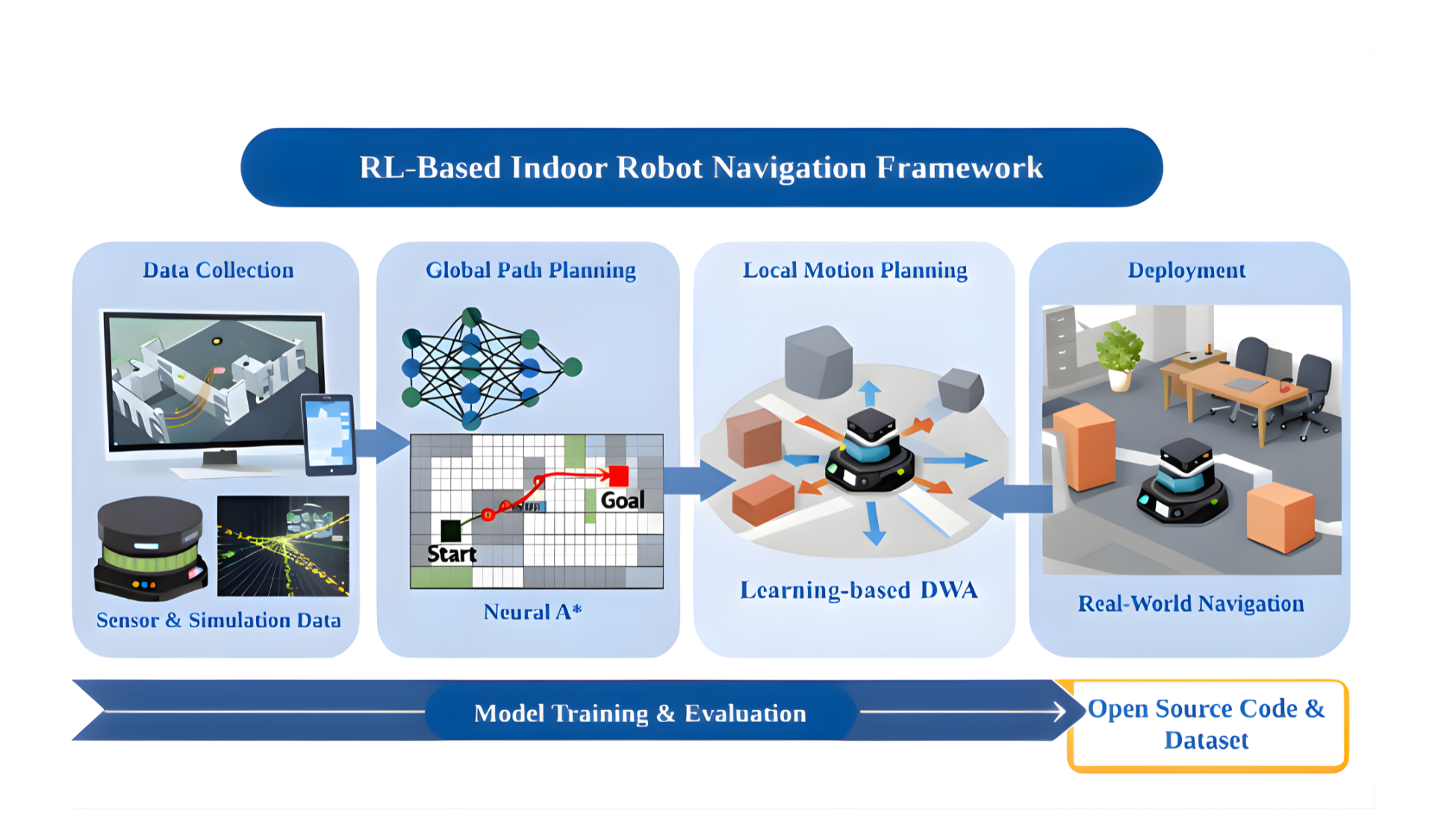}
    \caption{Overview of the proposed learning-based navigation framework. The pipeline includes data collection, global path planning, local motion planning, deployment, model training, and evaluation.}
    \label{fig:intro_framework}
\end{figure}

The main contributions of this paper are summarized as follows:
\begin{itemize}
    \item A hybrid learning-based navigation framework integrating a supervised neural global planner with a DWA-based local candidate selector.
    \item A two-stage local-planning policy trained by behavior cloning and PPO refinement with validity-aware action masking.
    \item Simulation and real-world validation demonstrating improved path quality and smoother motion compared with conventional DWA.
\end{itemize}

 \section{Related Work}
\subsection{Global Path Planning}

Global path planning provides map-level routes for mobile robot navigation. Classical methods mainly rely on graph-search or sampling-based algorithms, among which A* is widely used because of its efficiency and reliability on structured maps.

Recent learning-based planners, such as Neural A*~\cite{ruy2021neurala}, 
learning-guided heuristics~\cite{takahashi2019heuristic}, 
MPNet~\cite{qureshi2019mpnet}, 
and PRM-RL~\cite{faust2018prmrl}, 
have improved planning flexibility. In contrast to methods focusing mainly on global planning, this paper integrates a supervised neural global planner with a feasibility-preserving DWA-based local candidate selector for closed-loop indoor navigation.

\subsection{Local Path Planning}

Local path planning generates real-time motion commands for path tracking and obstacle avoidance. DWA is a representative velocity-space planner that selects admissible commands under dynamic constraints, obstacle clearance, and goal-progress criteria~\cite{fox1997dwa}. 
Although DWA has been extended in practical navigation systems~\cite{cai2022_jps_dwa}, 
it can still be conservative in cluttered environments.

Learning-based local planners have been studied to improve adaptability, including DWA-RL~\cite{patel2021dwarl}, 
RL-based navigation in crowded or cluttered scenes~\cite{kobayashi2023_qlearning_dwa,zhu2017targetdriven,long2018decentralizedrl,everett2021pedestrianrl,chen2020mapbasedrl,yao2021crowdaware}, 
velocity-obstacle-based DRL~\cite{xie2023drlvo}, 
RL-enhanced DWA~\cite{jiang2024improved}, 
and dimension-configurable local planning~\cite{zhang2025drldclp}. 
Practical mobile robot systems also emphasize the integration of planning, perception, and decision making for real-world deployment~\cite{lee2022odsbot}. 
Different from end-to-end policies that directly predict continuous commands, our method selects feasible candidates from the DWA action lattice, preserving the constraint-aware structure of DWA while improving motion smoothness and path quality through behavior cloning and PPO refinement.

    \section{Methodology}
    
    \subsection{System Overview}
    
\begin{figure}[htbp]
    	\centering
    	\includegraphics[width=0.46\textwidth]{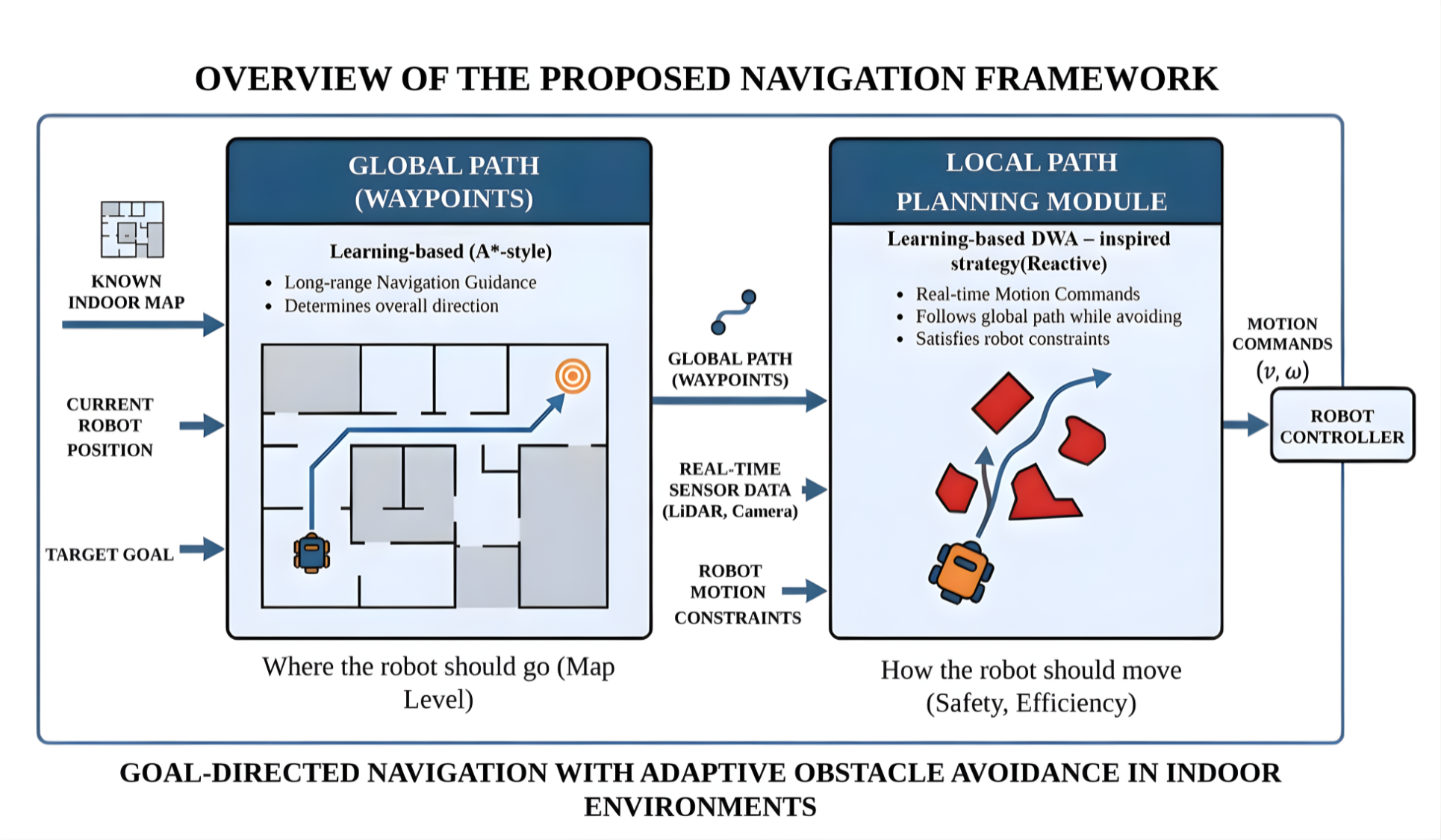}
    	\caption{Overview of the proposed navigation framework.}
    	\label{fig:system_overview}
    \end{figure}

    The proposed framework consists of a global planner and a local planner, as shown in Fig.~\ref{fig:system_overview}. The global planner generates a feasible route on a known indoor map, while the local planner produces real-time commands for path following and obstacle avoidance. The global planner is trained by supervised learning from expert demonstrations, and the local planner is initialized by behavior cloning and refined by reinforcement learning.
    
    The global planner follows a learning-based A*-inspired design, whereas the local planner is implemented as the proposed Learning-Based DWA, enabling closed-loop indoor navigation.
    
    \subsection{Robot Kinematic Model}
    
    The robot follows a differential-drive model with state $\mathbf{x}=[x,y,\theta]^T$~\cite{campion1996structural,thai2022ddmr}. The global planner provides a reference path, and the local planner generates feasible velocity commands $(v,\omega)$ for path following and obstacle avoidance.
    
    \subsection{Global Path Planning Module}
    
    The global planner generates a feasible map-level route on a known indoor costmap by learning a directional policy from trajectories produced by a cost-aware A* expert.
    
    \subsubsection{Cost-Aware A* Expert Planner}
    
    Training targets are generated by a cost-aware A* planner on a 2D costmap. The planner expands 8-connected neighbors and selects the node minimizing $f(n)=g(n)+h(n)$, where $g(n)$ is the accumulated cost and $h(n)$ is the Euclidean heuristic.

    Successor cost includes step, turning, and high-cost penalties. Unknown and occupied cells are excluded. The resulting paths provide state--action supervision.
    
    \subsubsection{Planning Representation and Network Architecture}
    
    At each planning step, the input is a five-channel tensor $\mathbf{x}_t \in \mathbb{R}^{192 \times 192 \times 5}$, encoding obstacle occupancy, normalized costmap value, unknown-space mask, current robot position, and goal position. The target output is the expert next-step action $a_t^E \in \{0,1,\dots,7\}$, corresponding to the 8-connected directions $\{\mathrm{N}, \mathrm{NE}, \mathrm{E}, \mathrm{SE}, \mathrm{S}, \mathrm{SW}, \mathrm{W}, \mathrm{NW}\}$. An auxiliary path mask is used for dense spatial supervision.
    
\begin{figure}[htbp]
    	\centering
    	\includegraphics[width=0.48\textwidth]{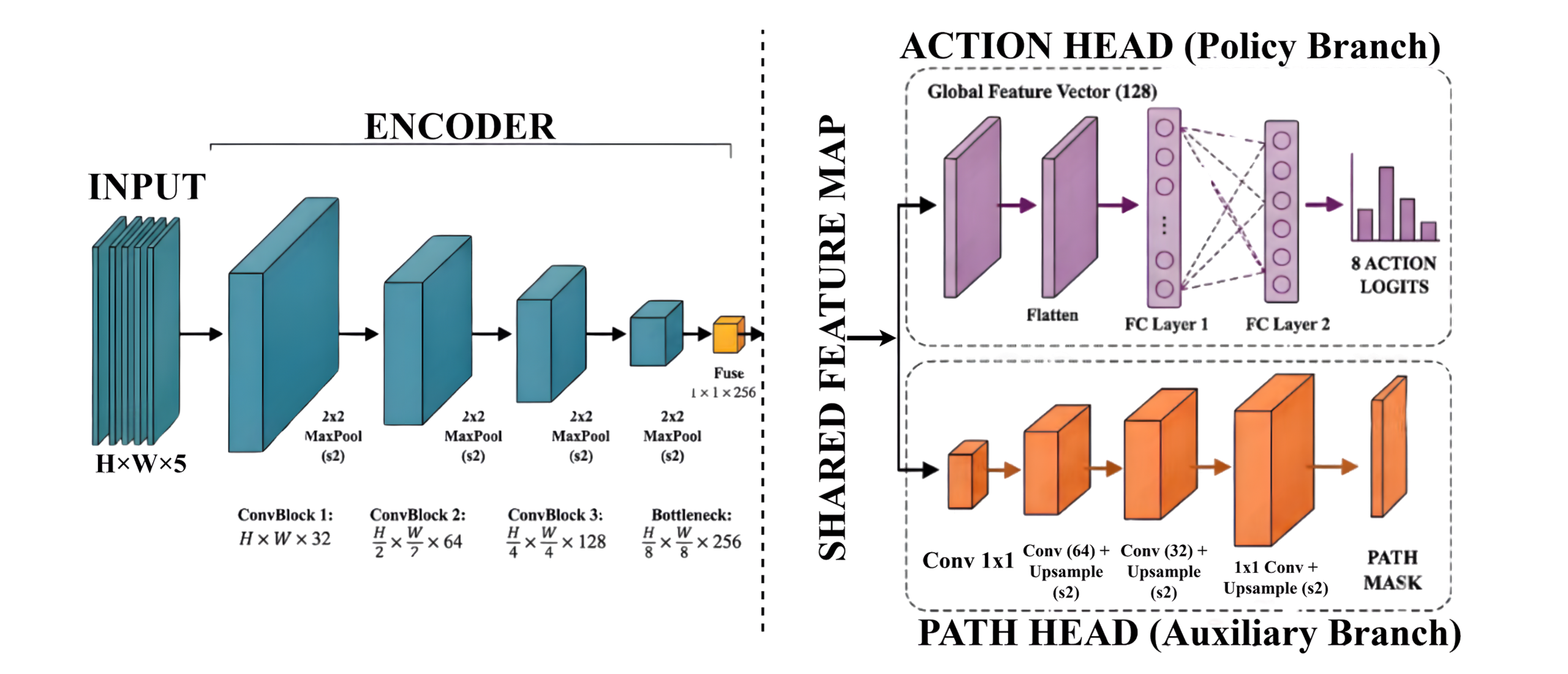}
    	\caption{Network architecture of the proposed neural global planner, consisting of an encoder, an action head, and an auxiliary path head.}
    	\label{fig:a_star_model}
    \end{figure}

    The global planner is implemented as a convolutional policy network, denoted by $\pi_{\theta}^{\mathrm{G}}$, which maps the input to logits over the eight candidate directions as $\mathbf{z}_t=\pi_{\theta}^{\mathrm{G}}(\mathbf{x}_t)\in\mathbb{R}^{8}$. Fig.~\ref{fig:a_star_model} shows the proposed network.
    
    The network uses a shared convolutional encoder and an action head, while the auxiliary path branch is used only during training.
    
    \subsubsection{Training Objective}
    
    Let the supervised training dataset be
    \begin{equation}
    	\mathcal{D}_{\mathrm{G}} = \{(\mathbf{x}_t, a_t^E, \mathbf{M}_t)\}_{t=1}^{M},
    \end{equation}
    where $\mathbf{x}_t$, $a_t^E$, and $\mathbf{M}_t$ denote the input tensor, expert action, and auxiliary path mask, respectively. The action-classification loss is
    \begin{equation}
    	\mathcal{L}_{\mathrm{act}}(\theta)
    	=
    	-\frac{1}{M}\sum_{t=1}^{M}\log \pi_{\theta}^{\mathrm{G}}(a_t^E \mid \mathbf{x}_t),
    \end{equation}
    and the auxiliary path loss is
    \begin{equation}
    	\mathcal{L}_{\mathrm{path}}(\theta)
    	=
    	\frac{1}{M}\sum_{t=1}^{M}\ell(\hat{\mathbf{M}}_t,\mathbf{M}_t),
    \end{equation}
    where $\hat{\mathbf{M}}_t$ is the predicted path mask. The total loss is
    \begin{equation}
    	\mathcal{L}_{\mathrm{G}}(\theta)
    	=
    	\mathcal{L}_{\mathrm{act}}(\theta)
    	+
    	\lambda_{\mathrm{path}}\mathcal{L}_{\mathrm{path}}(\theta),
    \end{equation}
    where $\lambda_{\mathrm{path}}$ controls the auxiliary-loss weight.
    
    \subsubsection{Runtime Rollout-Based Planning}
At deployment, planning proceeds autoregressively in grid space. From $\mathbf{p}_t$, the network predicts an action, masks invalid moves, and updates the state as $\mathbf{p}_{t+1}=\mathbf{p}_t+\Delta(a_t)$.

Rollout stops at the goal, on invalid progression, or at the horizon limit. The resulting path is then converted to world coordinates.
    
    \subsection{Local Path Planning Module}
    
    The local planner generates real-time velocity commands for tracking and obstacle avoidance under kinematic and dynamic constraints. Two variants are used: conventional DWA and a learning-based DWA candidate selector. The conventional DWA serves as both the rule-based baseline and the expert used for supervision.
    
    \subsubsection{Conventional DWA}
    
    At each control cycle, DWA samples admissible \((v,\omega)\) pairs within the dynamic window and rolls out short-horizon trajectories. Each candidate is scored by obstacle, goal, speed, and path-tracking costs. After normalization, the total cost is
\begin{equation}
	C_i^{\mathrm{total}}
	=
	\lambda_{\mathrm{obs}}\tilde{C}_i^{\mathrm{obs}}
	+\lambda_{\mathrm{goal}}\tilde{C}_i^{\mathrm{goal}}
	+\lambda_{\mathrm{speed}}\tilde{C}_i^{\mathrm{speed}}
	+\lambda_{\mathrm{path}}\tilde{C}_i^{\mathrm{path}}.
\end{equation}
The valid candidate with minimum cost is selected.
    
    \subsubsection{Learning-Based DWA Candidate Selection}
    
    Instead of learning continuous control, the proposed method performs discrete selection over the DWA action lattice.
    
    At time step \(t\), the observation contains a global context vector and candidate-wise features. The context vector is \(\mathbf{x}^{\mathrm{ctx}}_t=
    	[\Delta x_g,\ \Delta y_g,\ \Delta \theta_g,\ v_r,\ \omega_r,\ d_g,\ d_{\min}]\), where the terms denote goal position in the robot frame, goal heading error, current robot velocities, goal distance, and minimum frontal obstacle distance.
    
    For candidate \(i\), the feature vector is \(\mathbf{f}_t^{(i)}=
    	[v_i,\ \omega_i,\ c_i^{\mathrm{obs}},\ c_i^{\mathrm{goal}},\ c_i^{\mathrm{speed}},\ c_i^{\mathrm{path}},\ c_i^{\mathrm{total}},\ m_i]\), where \(m_i\in\{0,1\}\) is the validity flag. The complete observation is \(\mathbf{o}_t=
    	[\mathbf{x}^{\mathrm{ctx}}_t,\mathbf{f}_t^{(0)},\mathbf{f}_t^{(1)},\dots,\mathbf{f}_t^{(N-1)}]\).
    
    With \(N=63\) candidates and \(8\) features per candidate, the input dimension is \(d_o = 7 + 63\times 8 = 511\).
    
    The expert label is the candidate index chosen by conventional DWA: \(a_t^E \in \{0,\dots,N-1\}\).
    
    \begin{figure}[htbp]
    	\centering
    	\includegraphics[width=0.9\columnwidth]{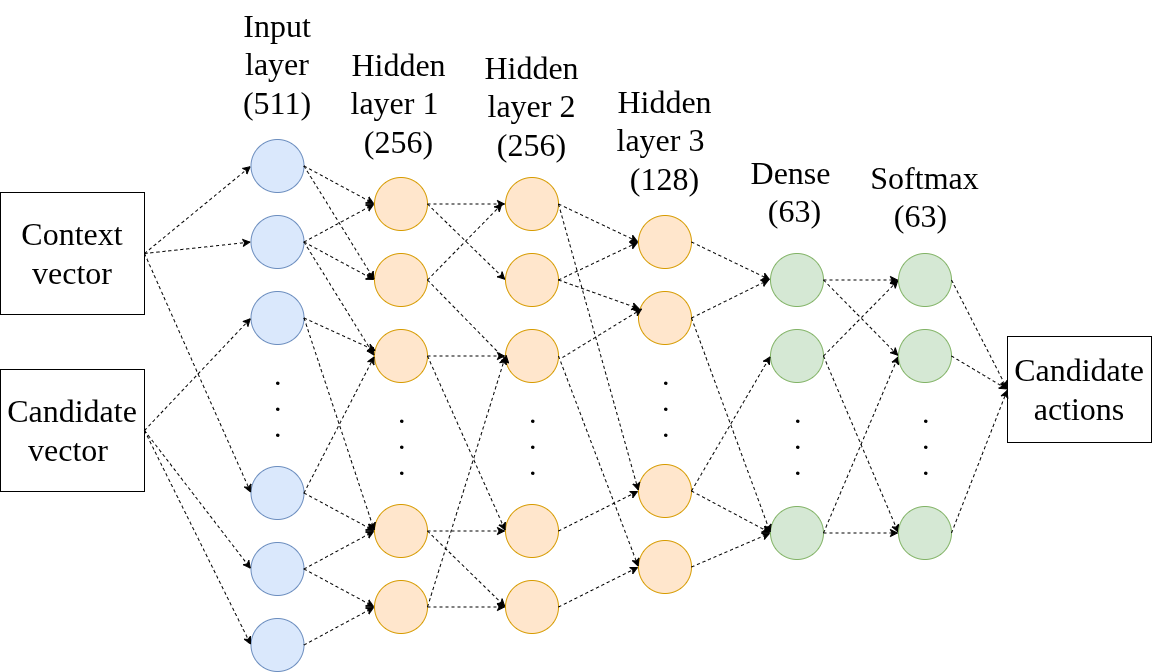}
    	\caption{MLP-based behavior-cloning architecture for DWA candidate selection from context and candidate vectors.}
    	\label{fig:bc_dwa_model}
    \end{figure}
    
    Each sample is a state-action pair \((\mathbf{o}_t,a_t^E)\). Let \(\pi_\theta(\mathbf{o}_t)\) denote the candidate-selection policy. The network outputs logits \(\mathbf{z}_t\in\mathbb{R}^{N}\), which define
    \begin{equation}
    	\pi_\theta(a=i\mid\mathbf{o}_t)=
    	\frac{\exp(z_{t,i})}{\sum_{j=0}^{N-1}\exp(z_{t,j})}.
    \end{equation}
    
    Given \(\mathcal{D}=\{(\mathbf{o}_t,a_t^E)\}_{t=1}^{M}\), behavior cloning minimizes
    \begin{equation}
    	\mathcal{L}_{\mathrm{BC}}(\theta)=
    	-\frac{1}{M}\sum_{t=1}^{M}\log \pi_\theta(a_t^E\mid\mathbf{o}_t).
    \end{equation}
    
    During inference, validity-aware logit adjustment is used:
    \begin{equation}
    	\tilde{z}_{t,i}=z_{t,i}+\beta(1-m_i),
    	\qquad \beta<0,
    \end{equation}
    and the deployed action is \(a_t^*=\arg\max_{i\in\{0,\dots,N-1\}}\tilde{z}_{t,i}\).
    
\subsubsection{PPO Fine-Tuning of the Candidate-Selection Policy}

\begin{figure}[htbp]
	\centering
	\includegraphics[width=0.9\columnwidth]{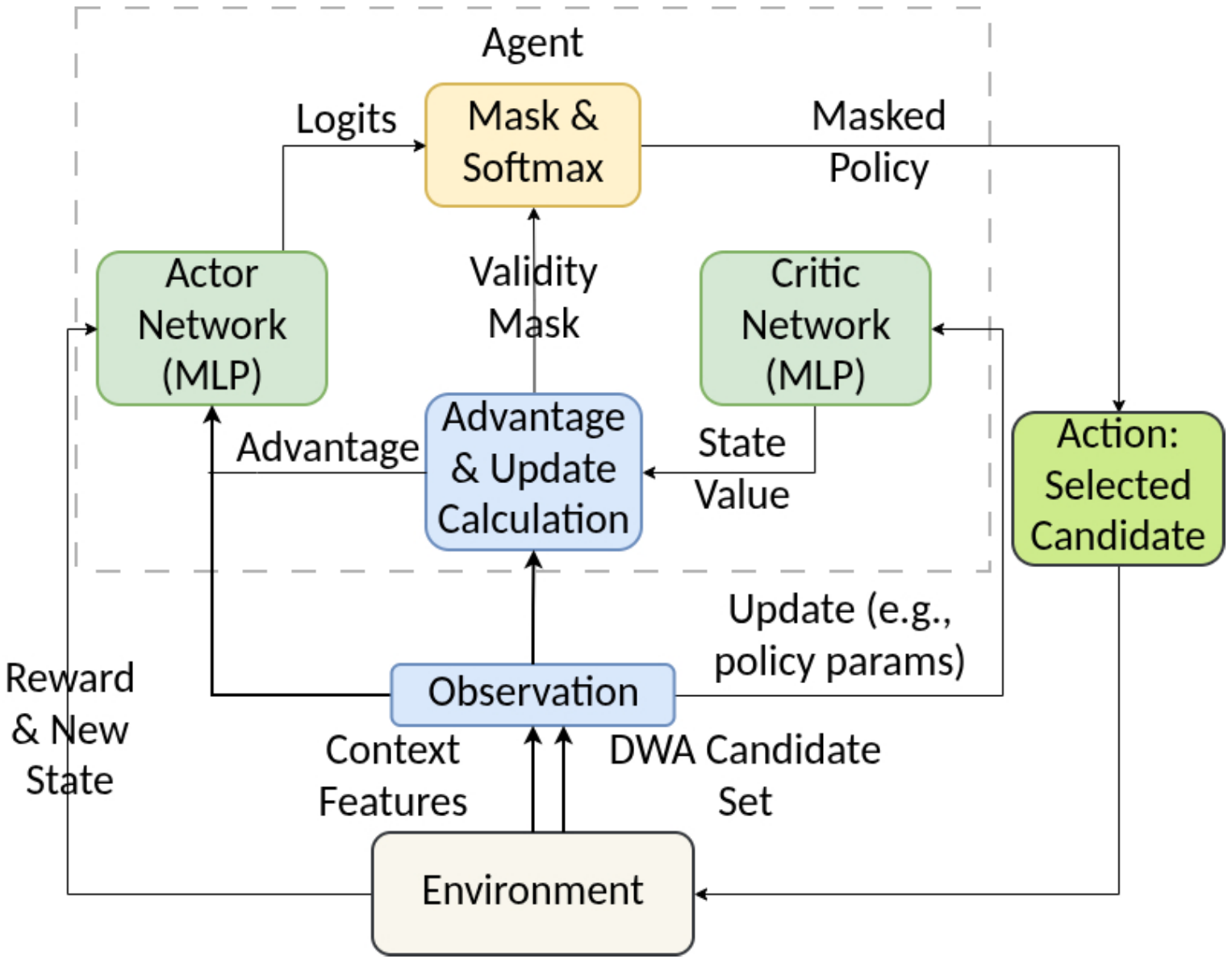}
	\caption{Overview of PPO fine-tuning for feasibility-aware DWA candidate selection.}
	\label{fig:ppo_framework}
\end{figure}

Behavior cloning provides the initial policy, and PPO is used for lightweight fine-tuning to improve robustness while preserving the same discrete DWA candidate space. Fig.~\ref{fig:ppo_framework} illustrates the PPO fine-tuning framework for feasibility-aware DWA candidate selection.

The same validity-aware logit adjustment is retained during PPO fine-tuning, so the policy remains constrained to feasible DWA candidates. The resulting policy is written as
\begin{equation}
	\pi_{\theta}(a=i\mid \mathbf{o}_t)=
	\frac{\exp(\tilde{z}_{t,i})}
	{\sum_{j=0}^{N-1}\exp(\tilde{z}_{t,j})}.
\end{equation}

The actor is initialized from the behavior-cloning model. PPO then maximizes the clipped surrogate objective
\begin{equation}
	J_{\mathrm{PPO}}(\theta)=
	\mathbb{E}_t
	\left[
	\min
	\left(
		\rho_t(\theta)\hat{A}_t,\,
		\mathrm{clip}(\rho_t(\theta),1-\epsilon,1+\epsilon)\hat{A}_t
	\right)
	\right].
\end{equation}

The reward promotes progress, safety, and smoothness, with large penalties for collision and timeout.
    \section{Experiments} 
\begin{figure}[htbp]
    	\centering
    	\includegraphics[width=0.35\textwidth]{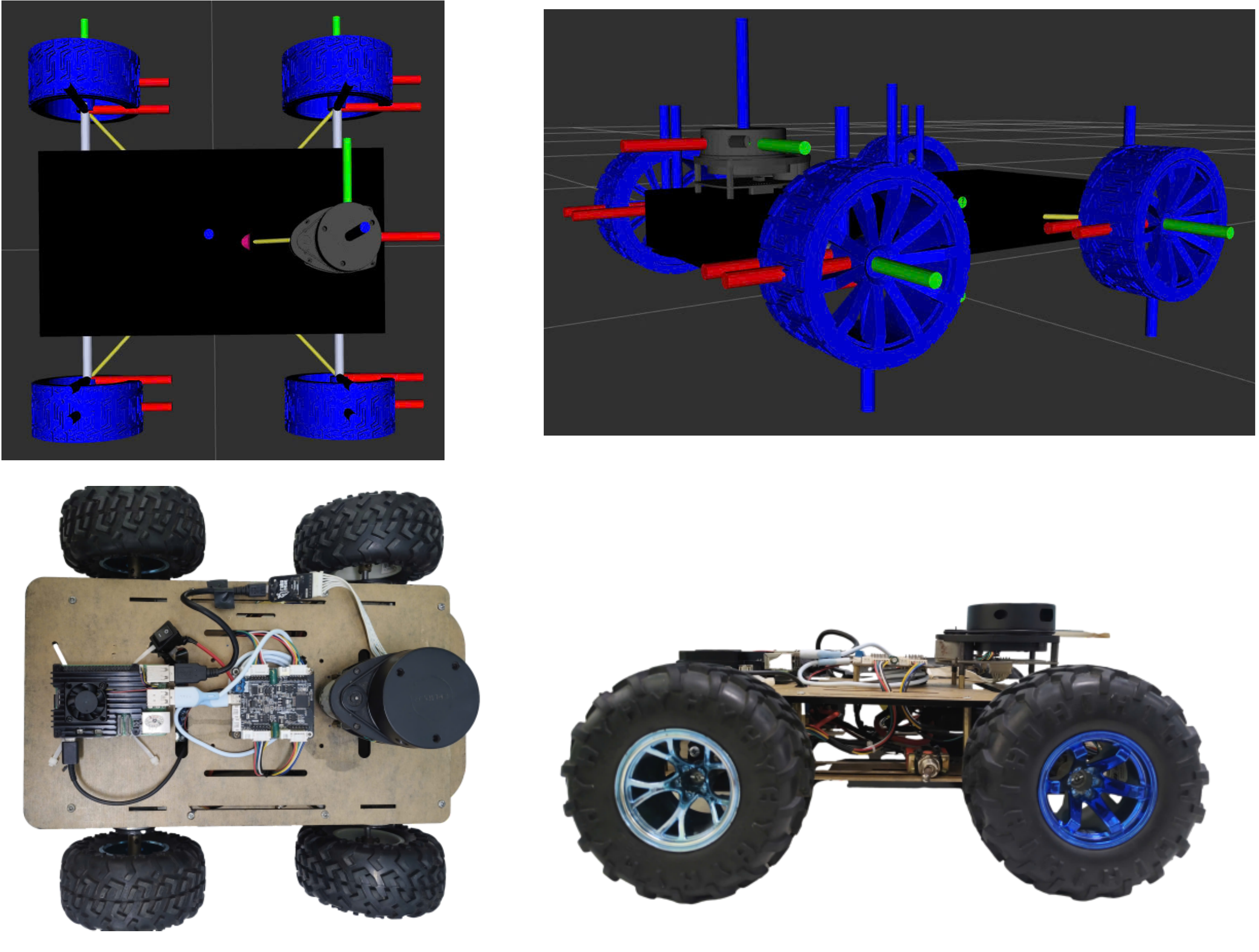}
        \caption{Experimental platforms: simulated robot and real robot equipped with DC motors, STM32 controller, 2D LiDAR, onboard ROS computer, and laptop-based learning module.}
    	\label{fig:sim_real_platform}
    \end{figure}

    Fig.~\ref{fig:sim_real_platform} shows the simulation model and the real robot used in the experiments.
    
    All learning-based experiments were conducted on a laptop with an Intel Core i5-12500H CPU and an NVIDIA GeForce RTX 3050 GPU.
    
    \subsection{Experiments on the Neural A* Global Planner}  
    \subsubsection{Dataset and Training Setup}

    The global-planning dataset contains 984 A*-expert episodes and 114,726 samples after state extraction and recovery-state augmentation. Inputs are cropped around the start--goal region and resized to $192\times192$.

Training uses batch size 64, learning rate $3\times10^{-4}$, weight decay $10^{-4}$, mixed precision, and standard augmentation.
        
    \subsubsection{Offline Evaluation}
    
    Table~\ref{tab:global_planner_results} summarizes the offline validation results. The best checkpoint is obtained at epoch 19, suggesting that one-step accuracy alone is insufficient for autoregressive planning.
    
    \begin{table}[htbp]
    	\centering
    	\refstepcounter{table}
    	\begin{center}
    	\footnotesize TABLE \Roman{table}\\[1mm]
        \footnotesize OFFLINE VALIDATION RESULTS OF THE PROPOSED\\
        \footnotesize GLOBAL PLANNER.
    	
        \end{center}
    	\label{tab:global_planner_results}
    	\begin{tabular}{lccc}
    		\toprule
    		\textbf{Checkpoint} & \textbf{Val. Loss} & \textbf{Action Acc.} & \textbf{Rollout SR} \\
    		\midrule
    		Best (Ep. 19)  & \textbf{0.5943} & \textbf{85.77\%} & \textbf{91.94\%} \\
    		Final (Ep. 25) & 0.6204          & 84.67\%          & 85.83\% \\
    		\bottomrule
    	\end{tabular}
    \end{table}
    
    \subsubsection{Runtime Comparison}
    
    Table~\ref{tab:planner_compare_100} reports the average runtime over 100 paired trials. GPU-based Neural A* remains reasonably close to A*, whereas CPU inference is impractical for deployment.
    
    \begin{table}[htbp]
    	\centering
    	\refstepcounter{table}
    	\begin{center}
           \footnotesize TABLE \Roman{table}\\[1mm]
           \footnotesize AVERAGE RUNTIME COMPARISON AMONG A*, NEURAL A* (GPU), AND NEURAL A* (CPU).\\
           
        \end{center}
    	\label{tab:planner_compare_100}
    	\begin{tabular}{lc}
    		\toprule
    		\textbf{Method} & \textbf{Avg. Runtime (ms)} \\
    		\midrule
    		A*              & \textbf{67.962} \\
    		Neural A* (GPU) & 80.784 \\
    		Neural A* (CPU) & 12643 \\
    		\bottomrule
    	\end{tabular}
    \end{table}
    
    Tables~\ref{tab:env_path_compare} and~\ref{tab:env_runtime_compare} further compare open and obstacle-rich environments. Neural A* shows stronger path quality in both settings, while runtime varies with environment structure.
    
    \begin{table}[htbp]
    	\centering
    	\refstepcounter{table}
    	\begin{center}
    	   \footnotesize TABLE \Roman{table}\\[1mm]
    	   \footnotesize PATH WIN RATES IN OPEN AND OBSTACLE-RICH ENVIRONMENTS.\\
    	   
        \end{center}
    	\label{tab:env_path_compare}
    	\begin{tabular}{lcc}
    		\toprule
    		\textbf{Planner} & \textbf{Open} & \textbf{Obstacle-Rich} \\
    		\midrule
    		A*        & 40.00\%          & 33.00\%          \\
    		Neural A* & \textbf{60.00\%} & \textbf{67.00\%} \\
    		\bottomrule
    	\end{tabular}
    \end{table}
    
    \begin{table}[htbp]
    	\centering
    	\refstepcounter{table}
    	\begin{center}
        	\footnotesize TABLE \Roman{table}\\[1mm]
        	\footnotesize RUNTIME WIN RATES IN OPEN AND OBSTACLE-RICH ENVIRONMENTS\\
        	\footnotesize UNDER THE GPU-BASED NEURAL A* CONFIGURATION.\\
        	
        \end{center}
    	\label{tab:env_runtime_compare}
    	\begin{tabular}{lcc}
    		\toprule
    		\textbf{Planner} & \textbf{Open} & \textbf{Obstacle-Rich} \\
    		\midrule
    		A*        & \textbf{72.00\%} & 46.00\%          \\
    		Neural A* & 28.00\%          & \textbf{54.00\%} \\
    		\bottomrule
    	\end{tabular}
    \end{table}

  \subsubsection{Representative Examples}

Table~\ref{tab:two_example_summary} summarizes the path length and runtime of two representative successful cases, and Fig.~\ref{fig:two_examples} visualizes the corresponding planning results.

\begin{table}[htbp]
	\centering
	\refstepcounter{table}
	\begin{center}
		\footnotesize TABLE \Roman{table}\\[1mm]
		\footnotesize PATH LENGTH AND RUNTIME OF TWO REPRESENTATIVE EXAMPLES.\\
		\footnotesize FOR EACH ENTRY, RESULTS ARE REPORTED AS A* / NEURAL A*.
	\end{center}
	\label{tab:two_example_summary}
	\begin{tabular}{lcc}
		\toprule
		\textbf{Metric} & \textbf{Example 1} & \textbf{Example 2} \\
		\midrule
		Path (m)  & 5.149 / \textbf{5.110} & 5.797 / \textbf{5.651} \\
		Time (ms) & \textbf{59.695} / 63.634 & 87.205 / \textbf{73.907} \\
		\bottomrule
	\end{tabular}
\end{table}

\refstepcounter{figure}
\begin{center}
	\begin{minipage}[t]{0.44\columnwidth}
		\centering
		\includegraphics[width=\linewidth]{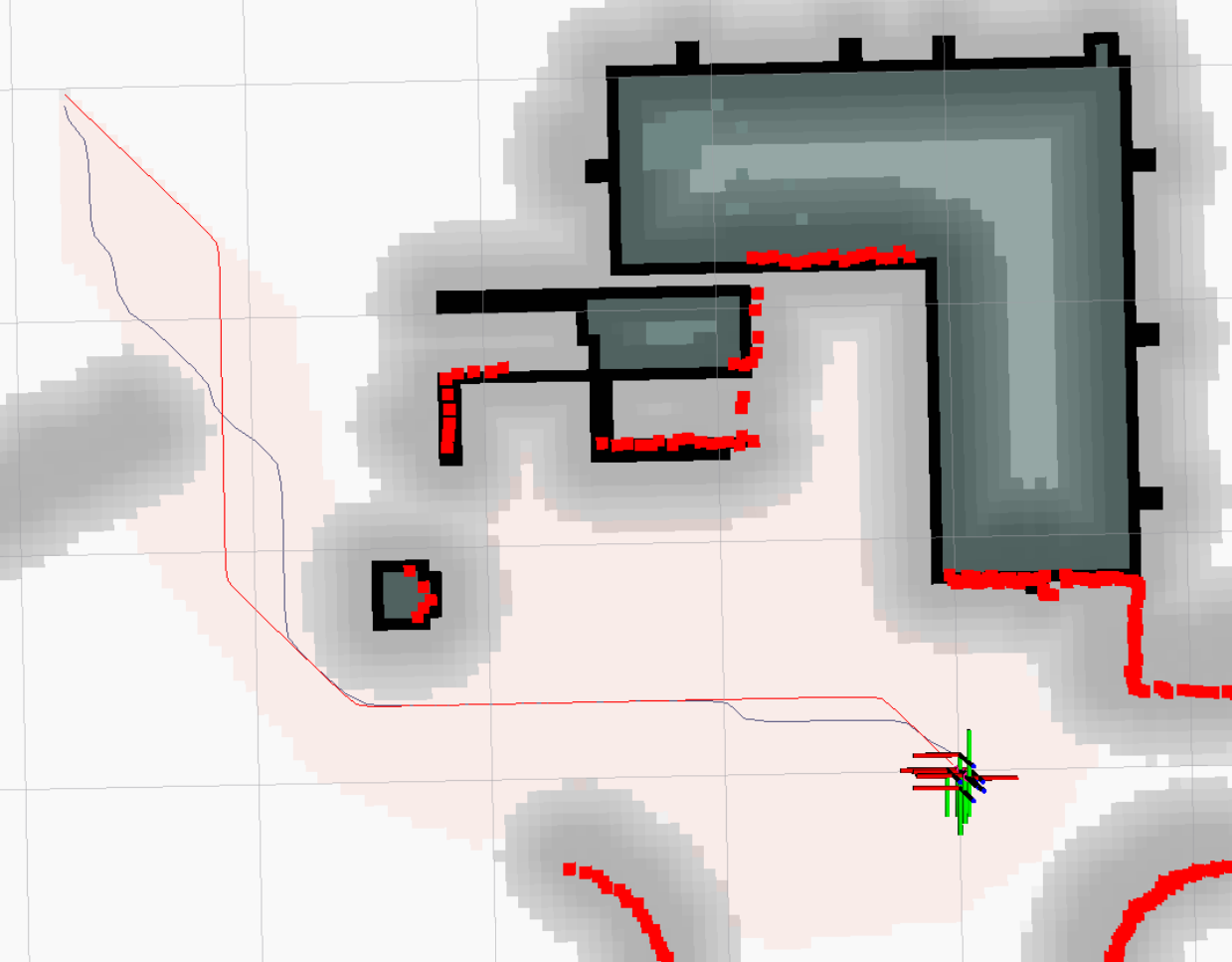}
		
		(a) Example 1.
	\end{minipage}\hfill
	\begin{minipage}[t]{0.44\columnwidth}
		\centering
		\includegraphics[width=\linewidth]{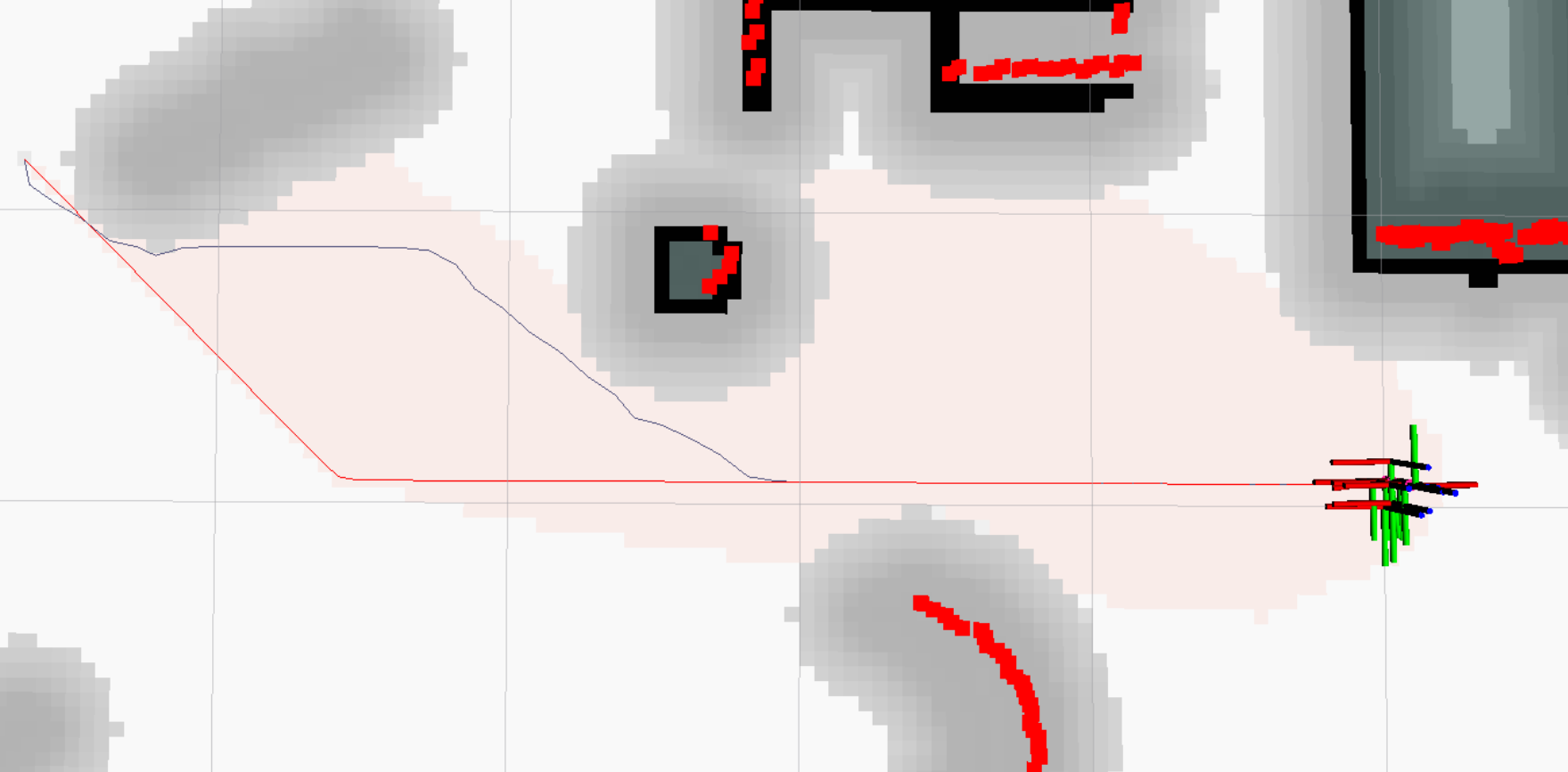}
		
		(b) Example 2.
	\end{minipage}
	
	\vspace{1mm}
	
	\footnotesize Fig. \thefigure: Representative deployment examples of the proposed neural planner compared with A*. The blue path denotes Neural A*, whereas the red path denotes classical A*.
	\label{fig:two_examples}
\end{center}

\subsection{Experiments on the Learning-Based DWA}
    
    \subsubsection{Training Procedure and Offline Comparison}
    \label{subsec:training}
    
    The local policy is trained from 38,512 valid expert-labeled samples collected over 74 navigation episodes using a two-stage pipeline: behavior cloning for initialization and PPO refinement on the same discrete DWA candidate space, with the actor initialized from BC and the same validity-aware action adjustment applied during training and deployment.
    
    Table~\ref{tab:offline_bc_ppo} reports the offline comparison and confirms the benefit of PPO refinement over the BC baseline.
    
    \begin{table}[htbp]
    	\centering
    	\refstepcounter{table}
    	\begin{center}
        		\footnotesize TABLE \Roman{table}\\[1mm]
        \footnotesize OFFLINE PERFORMANCE COMPARISON BETWEEN\\
        \footnotesize BC AND PPO-REFINED.
    		
    	\end{center}
    	\label{tab:offline_bc_ppo}
    	\begin{tabular}{lcc}
    		\toprule
    		\textbf{Metric} & \textbf{BC (Baseline)} & \textbf{PPO-Refined} \\
    		\midrule
    		Training Strategy & Supervised Imitation & BC + PPO Refinement \\
    		Accuracy          & 0.4233               & \textbf{0.5287} \\
    		Loss              & 2.0025               & \textbf{1.6013} \\
    		Meta Val. Acc.    & 0.3919               & \textbf{0.4721} \\
    		\bottomrule
    	\end{tabular}
    \end{table}
    
    \subsubsection{Performance in a Static Scenario}
    \label{subsec:static_case}
    
    The policy is evaluated in a static environment without dynamic obstacles over 10 runs. Table~\ref{tab:static_case_results} indicates a trade-off between path-following quality and traversal efficiency.
    
    \begin{table}[htbp]
    	\centering
    	\refstepcounter{table}
    	\begin{center}
        	\footnotesize TABLE \Roman{table}\\[1mm]
        \footnotesize AVERAGE PERFORMANCE IN THE STATIC SCENARIO\\
        \footnotesize OVER 10 RUNS.
        	
        \end{center}
    	\label{tab:static_case_results}
    	\begin{tabular}{lcc}
    		\toprule
    		\textbf{Metric} & \textbf{Learning-Based DWA} & \textbf{DWA} \\
    		\midrule
    		Path Length (m)    & \textbf{6.2404} & 6.3721 \\
    		Time to Goal (s)   & 36.0171          & \textbf{31.8045} \\
    		Tracking RMSE (m)  & \textbf{0.0633} & 0.1044 \\
    		RMS Jerk Linear    & \textbf{0.8230} & 1.7673 \\
    		RMS Jerk Angular   & \textbf{2.8796} & 3.0971 \\
    		\bottomrule
    	\end{tabular}
    \end{table}
    
    \begin{figure}[htbp]
    	\centering
    	\subfloat[Trajectory comparison]{\includegraphics[width=0.48\columnwidth]{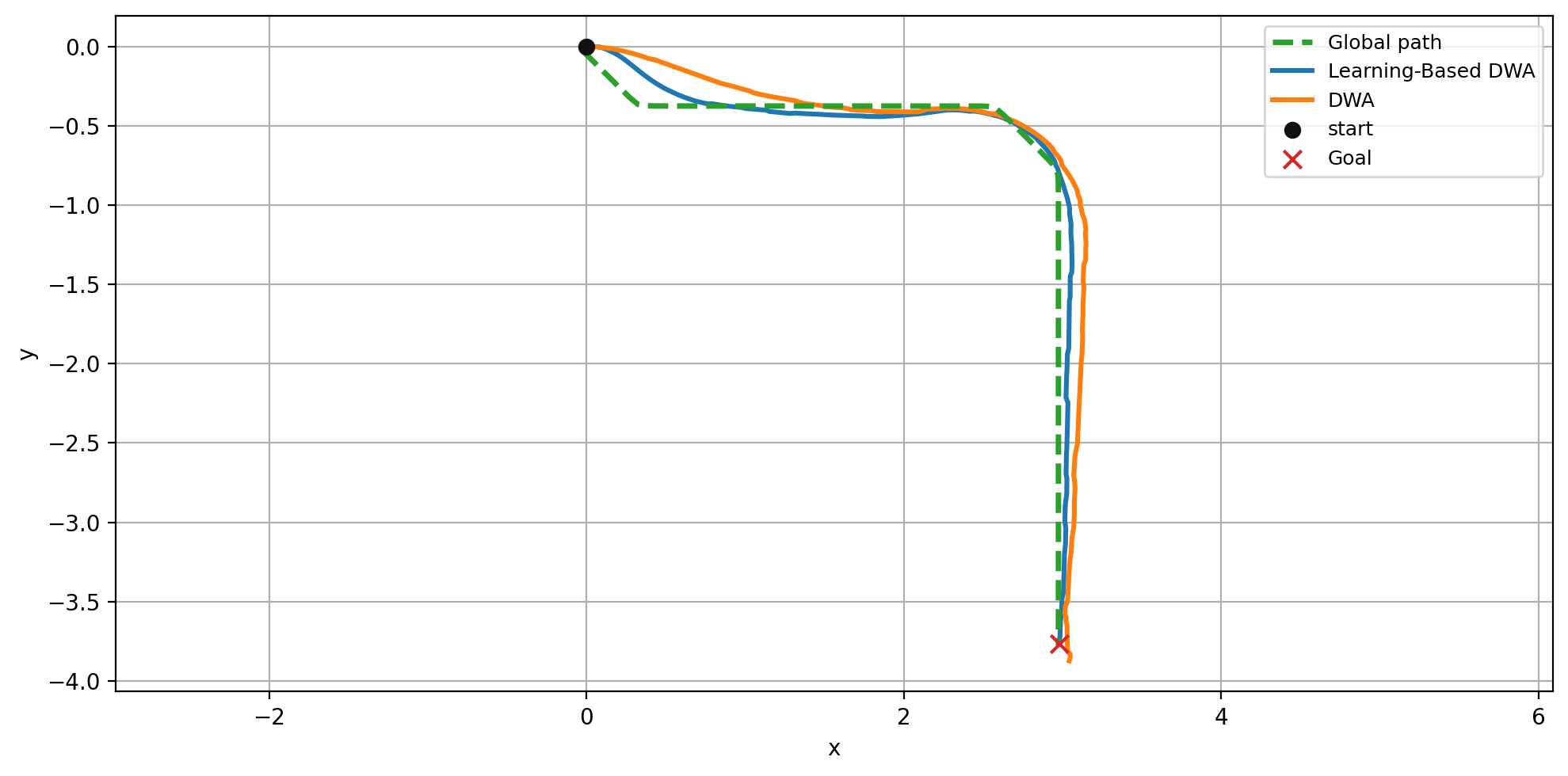}}
    	\hfill
    	\subfloat[Linear velocity profile]{\includegraphics[width=0.48\columnwidth]{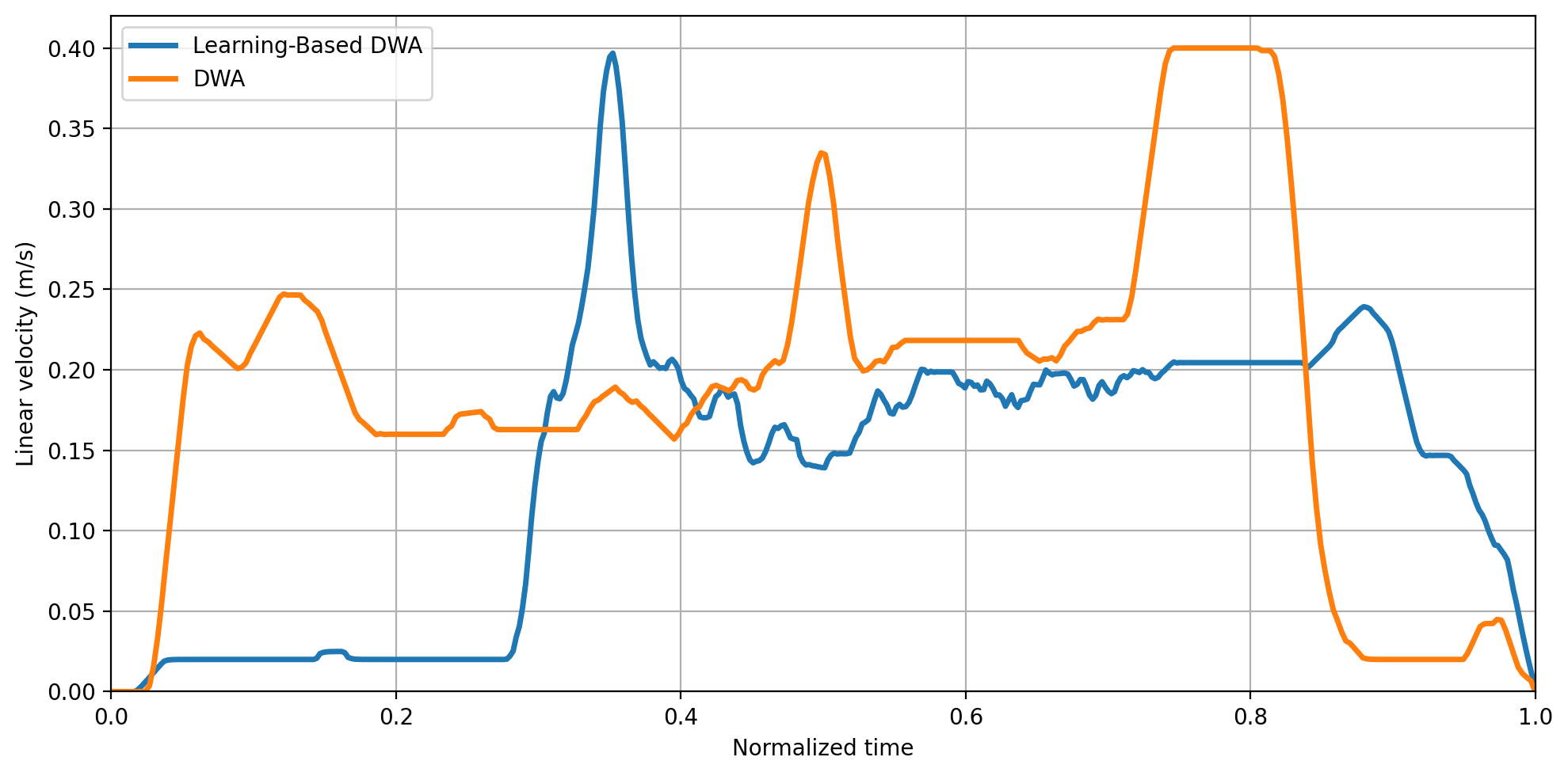}}\\
    	\subfloat[Tracking error profile]{\includegraphics[width=0.48\columnwidth]{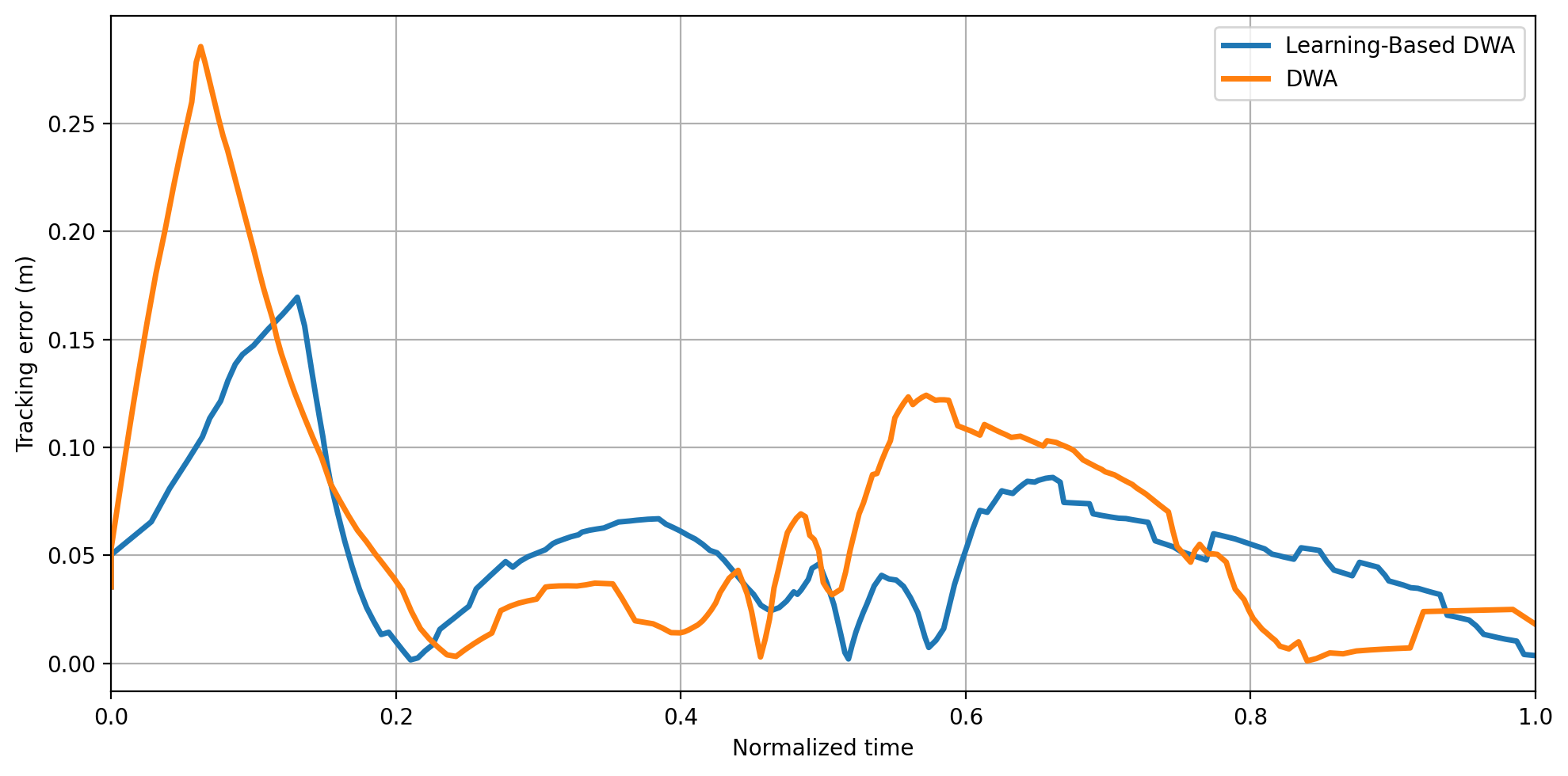}}
    	\hfill
    	\subfloat[Angular velocity profile]{\includegraphics[width=0.48\columnwidth]{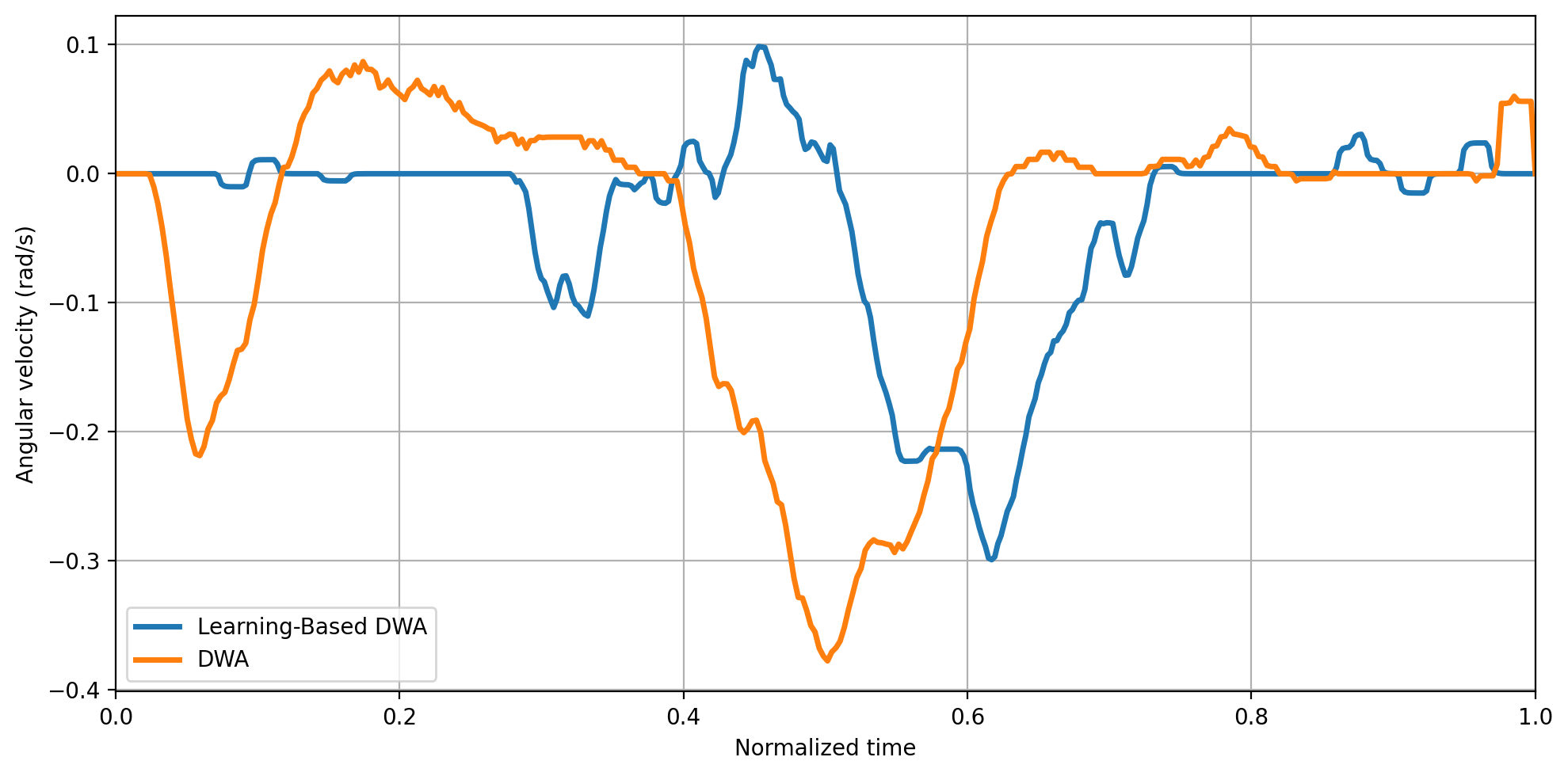}}
    	\caption{Qualitative comparison in the static scenario. The Learning-Based DWA trajectory and profiles are shown in blue, whereas those of the conventional DWA are shown in orange. The Learning-Based DWA achieves better path adherence and lower tracking error, while the conventional DWA reaches the goal faster.}
    	\label{fig:static_case}
    \end{figure}
    
    \subsubsection{Performance in an Obstacle Scenario}
    \label{subsec:obs_case}
    
    The policy is further evaluated in a cluttered obstacle scenario over 10 runs. Table~\ref{tab:obs_case_results} suggests more anticipative avoidance by the learned policy.
    \begin{table}[htbp]
    	\centering
    	\refstepcounter{table}
    	\begin{center}
    		\footnotesize TABLE \Roman{table}\\[1mm]
    		\footnotesize AVERAGE PERFORMANCE IN THE OBSTACLE SCENARIO\\  \footnotesize OVER 10 RUNS.
    	\end{center}
    	\label{tab:obs_case_results}
    	\begin{tabular}{lcc}
    		\toprule
    		\textbf{Metric} & \textbf{Learning-Based DWA} & \textbf{DWA} \\
    		\midrule
    		Path Length (m)    & \textbf{8.1256} & 8.3947 \\
    		Time to Goal (s)   & 43.0041          & \textbf{42.6190} \\
    		Tracking RMSE (m)  & 0.2363           & \textbf{0.2165} \\
    		RMS Jerk Linear    & \textbf{1.5157} & 1.5258 \\
    		RMS Jerk Angular   & \textbf{4.8440} & 6.1474 \\
    		\bottomrule
    	\end{tabular}
    \end{table}
    \begin{figure}[htbp]
    	\centering
    	\subfloat[Trajectory comparison]{\includegraphics[width=0.48\columnwidth]{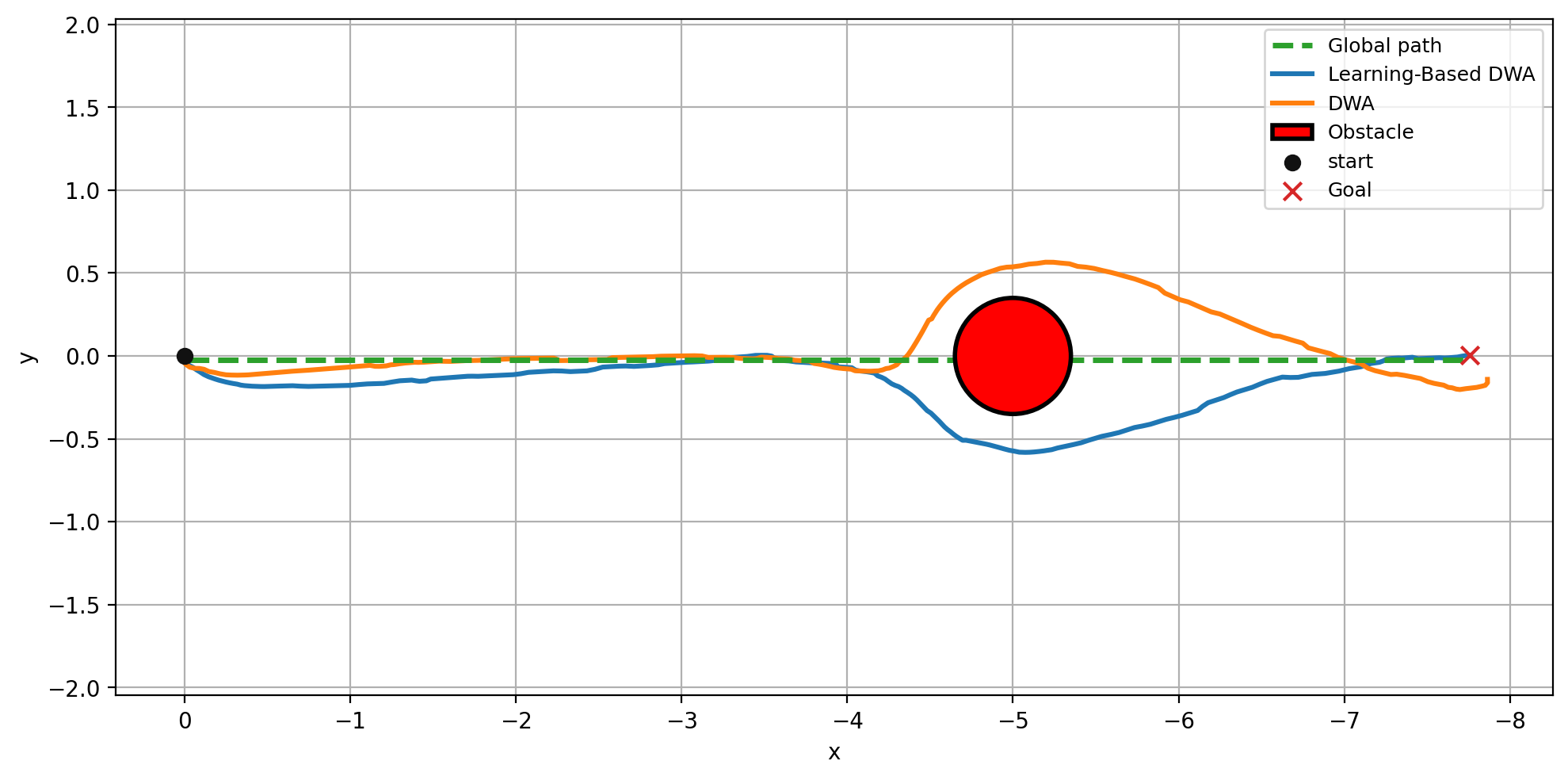}}
    	\hfill
    	\subfloat[Linear velocity profile]{\includegraphics[width=0.48\columnwidth]{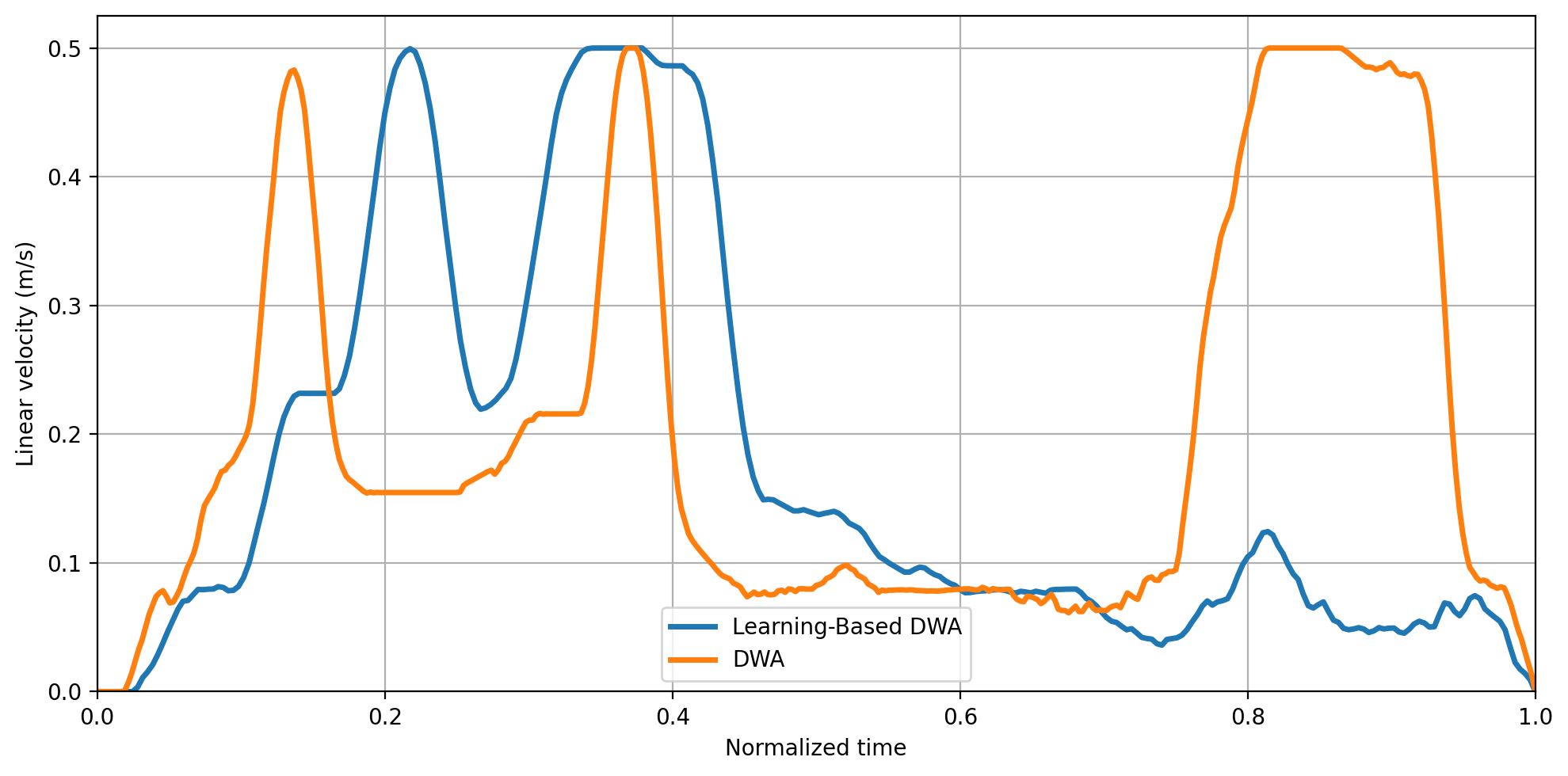}}\\
    	\subfloat[Tracking error profile]{\includegraphics[width=0.48\columnwidth]{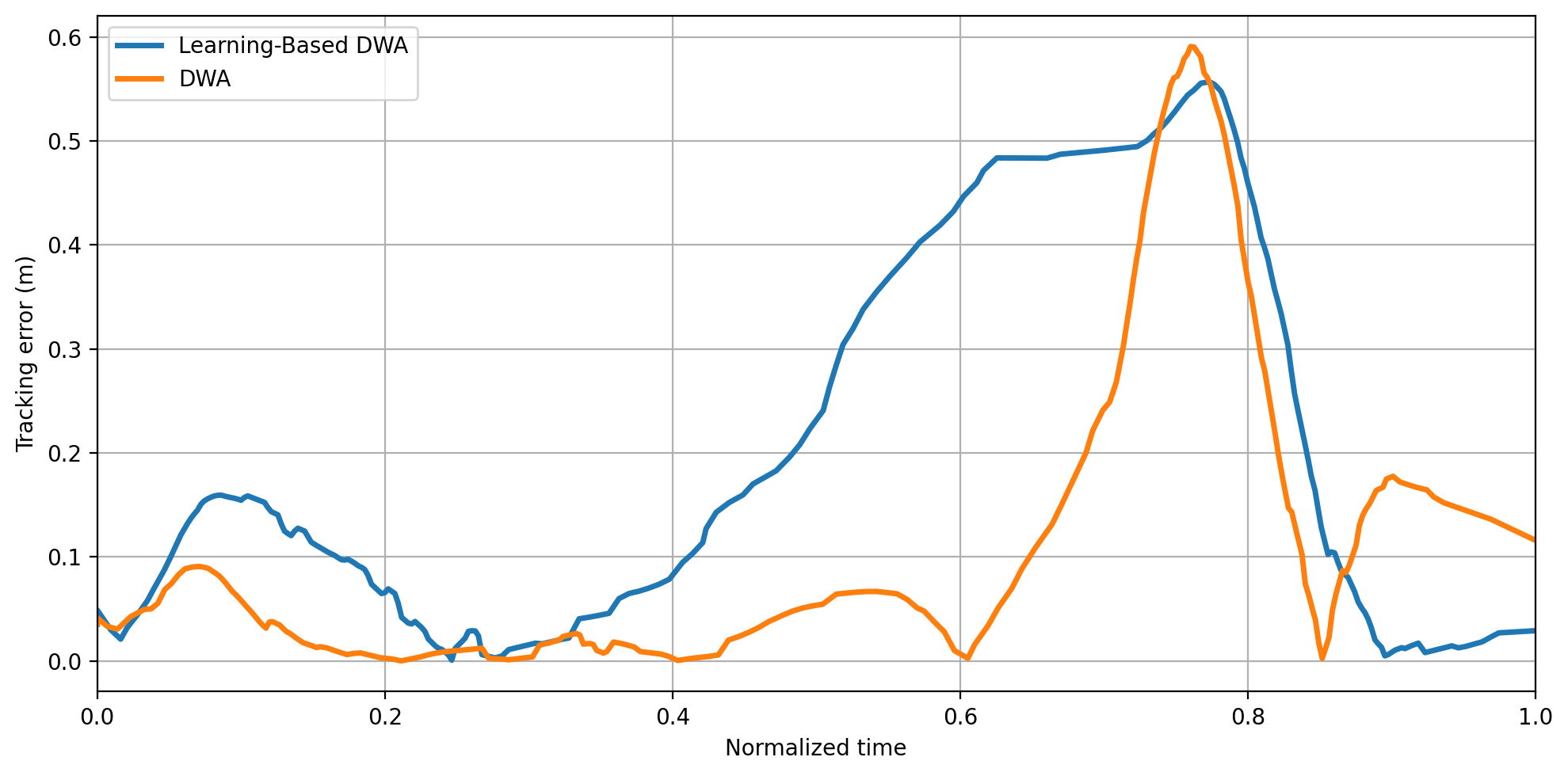}}
    	\hfill
    	\subfloat[Angular velocity profile]{\includegraphics[width=0.48\columnwidth]{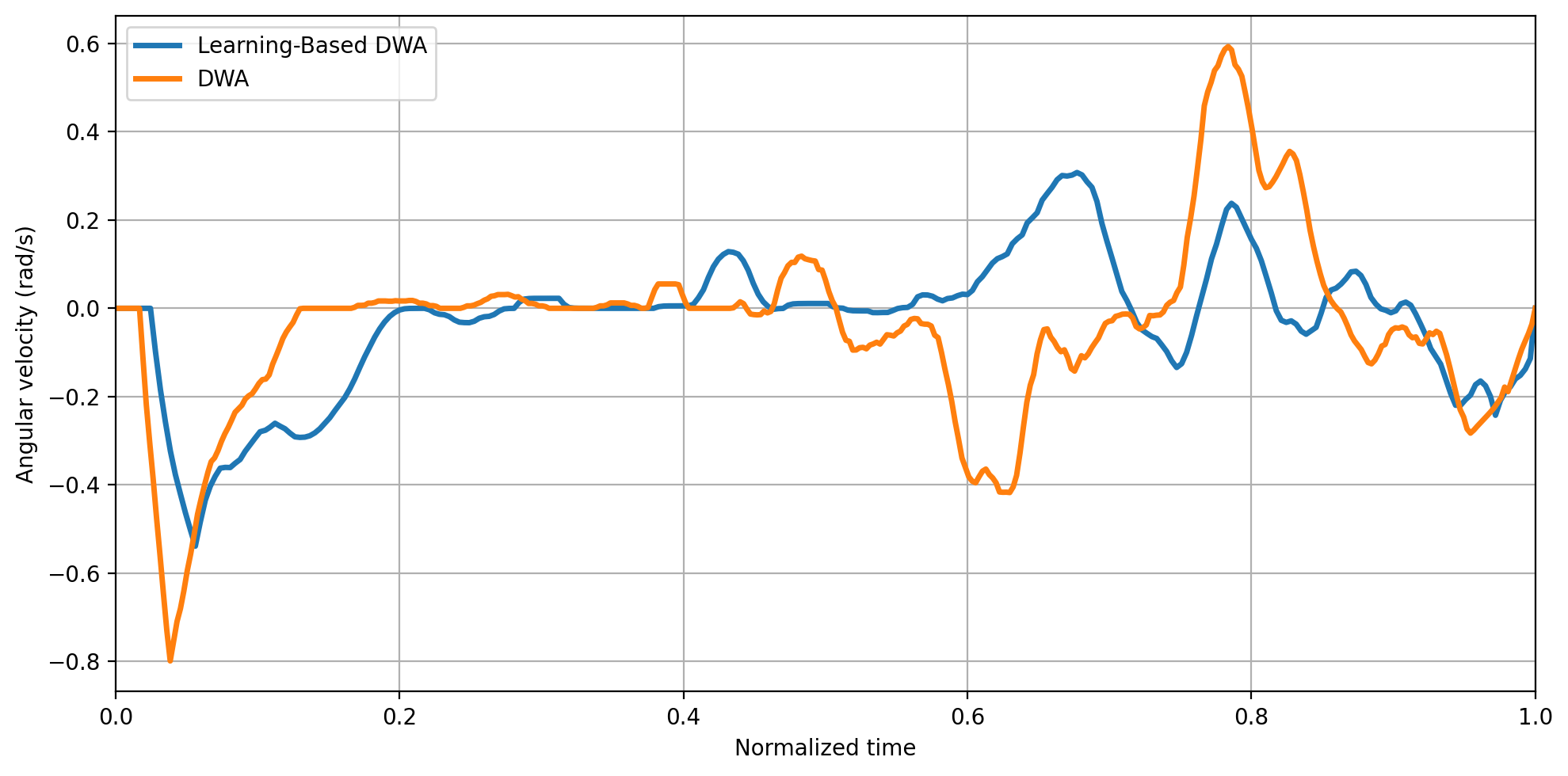}}
    	\caption{Qualitative comparison in the obstacle scenario: Learning-Based DWA (blue) and conventional DWA (orange). The Learning-Based DWA initiates obstacle avoidance earlier, leading to a slightly larger tracking error but a shorter path and smoother angular motion.}
    	\label{fig:obs_case}
    \end{figure}
    
    \subsubsection{Simulation Results and Closed-Loop Demonstration of the Local Planner}
    \label{subsec:simulation_results}
    
    The conventional DWA  and the proposed Learning-Based DWA are compared over 10 simulation runs. Tables~\ref{tab:variant_compare} and~\ref{tab:local_path_metrics} summarize the navigation and closed-loop tracking results.

    \begin{table}[htbp]
    	\centering
    	\refstepcounter{table}
    	\begin{center}
        	\footnotesize TABLE \Roman{table}\\[1mm]
        	\footnotesize COMPARISON BETWEEN CONVENTIONAL DWA AND LEARNING-BASED DWA IN SIMULATION.\\
        \end{center}
    	\label{tab:variant_compare}
        
    	\begin{tabular}{lcc}
    		\toprule
    		\textbf{Metric} & \textbf{Learning-Based DWA} & \textbf{DWA} \\
    		\midrule
    		Success Rate (\%)    & \textbf{100}    & \textbf{100} \\
    		Path Length (m)      & \textbf{6.5423} & 6.6391 \\
    		Time to Goal (s)     & 36.088          & \textbf{31.339} \\
    		Final Goal Error (m) & \textbf{0.1763} & 0.2983 \\
    		Avg. Speed (m/s)     & 0.1945          & \textbf{0.2190} \\
    		Linear Jerk          & \textbf{4.3950} & 4.9028 \\
    		Angular Jerk         & \textbf{14.6430} & 16.2486 \\
    		\bottomrule
    	\end{tabular}
    \end{table}
    The longer traversal time of the Learning-Based DWA reflects a trade-off between smoothness and efficiency. PPO refinement promotes stable candidate selection and reduces abrupt velocity changes, but makes the policy more conservative than conventional DWA.
    \begin{table}[htbp]
    	\centering
    	\refstepcounter{table}
    	\begin{center}
    		\footnotesize TABLE \Roman{table}\\[1mm]
    		\footnotesize CLOSED-LOOP TRACKING CONSISTENCY OF THE LEARNING-BASED DWA IN SIMULATION.\\
    	\end{center}
    	\label{tab:local_path_metrics}
    	\begin{tabular}{lc}
    		\toprule
    		\textbf{Metric} & \textbf{Value} \\
    		\midrule
    		Mean Cross-Track Error (odom-ref) (m) & 0.1490 \\
    		Command-to-Odometry RMSE (m)          & 0.2633 \\
    		\bottomrule
    	\end{tabular}
    \end{table}

    \begin{figure}[htbp]
    	\centering
    	\begin{subfigure}[b]{0.48\columnwidth}
    		\centering
    		\includegraphics[width=\linewidth]{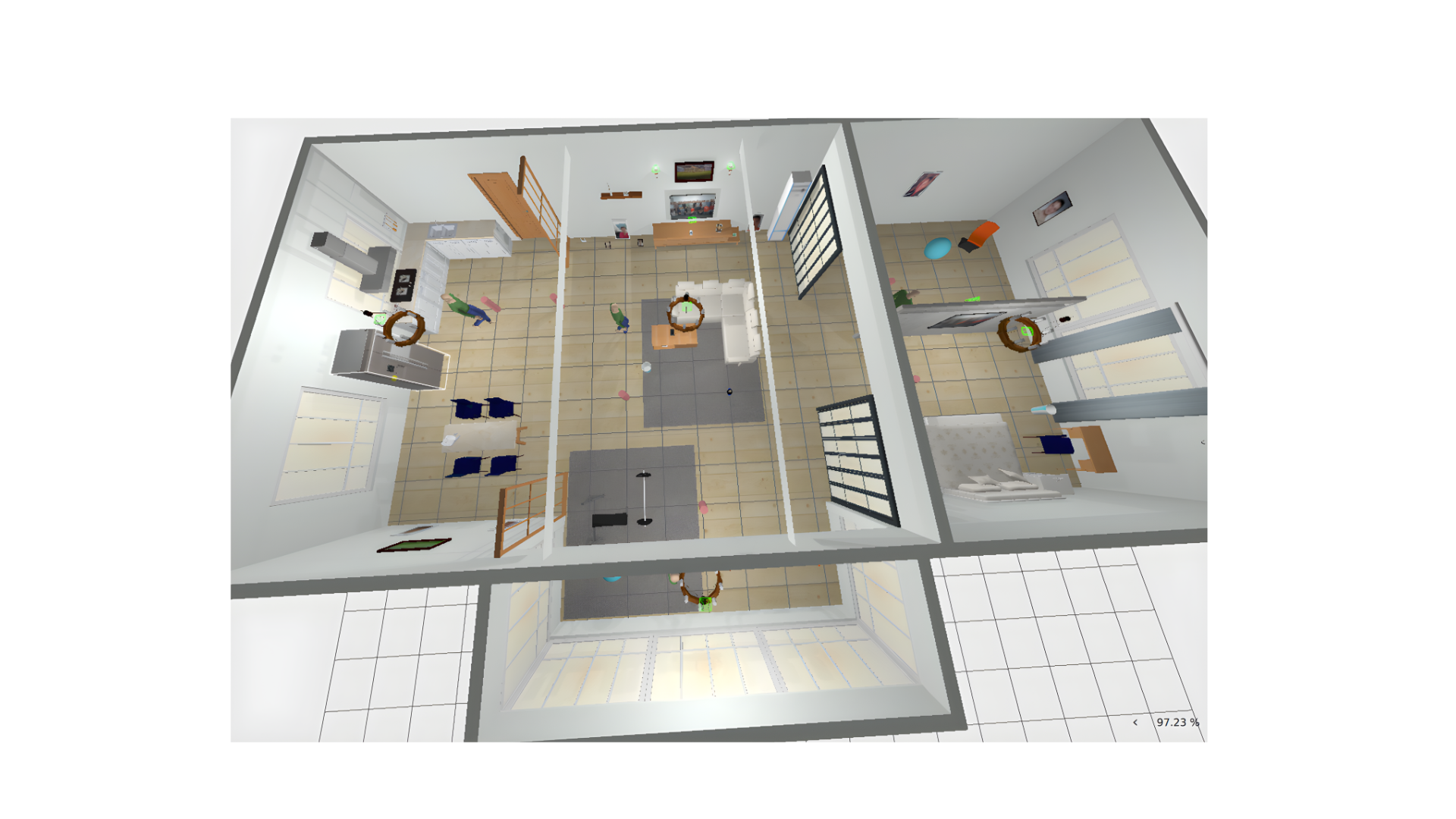}
    		\caption{Gazebo view}
    		\label{fig:gazebo}
    	\end{subfigure}
    	\hfill
    	\begin{subfigure}[b]{0.48\columnwidth}
    		\centering
    		\includegraphics[width=\linewidth]{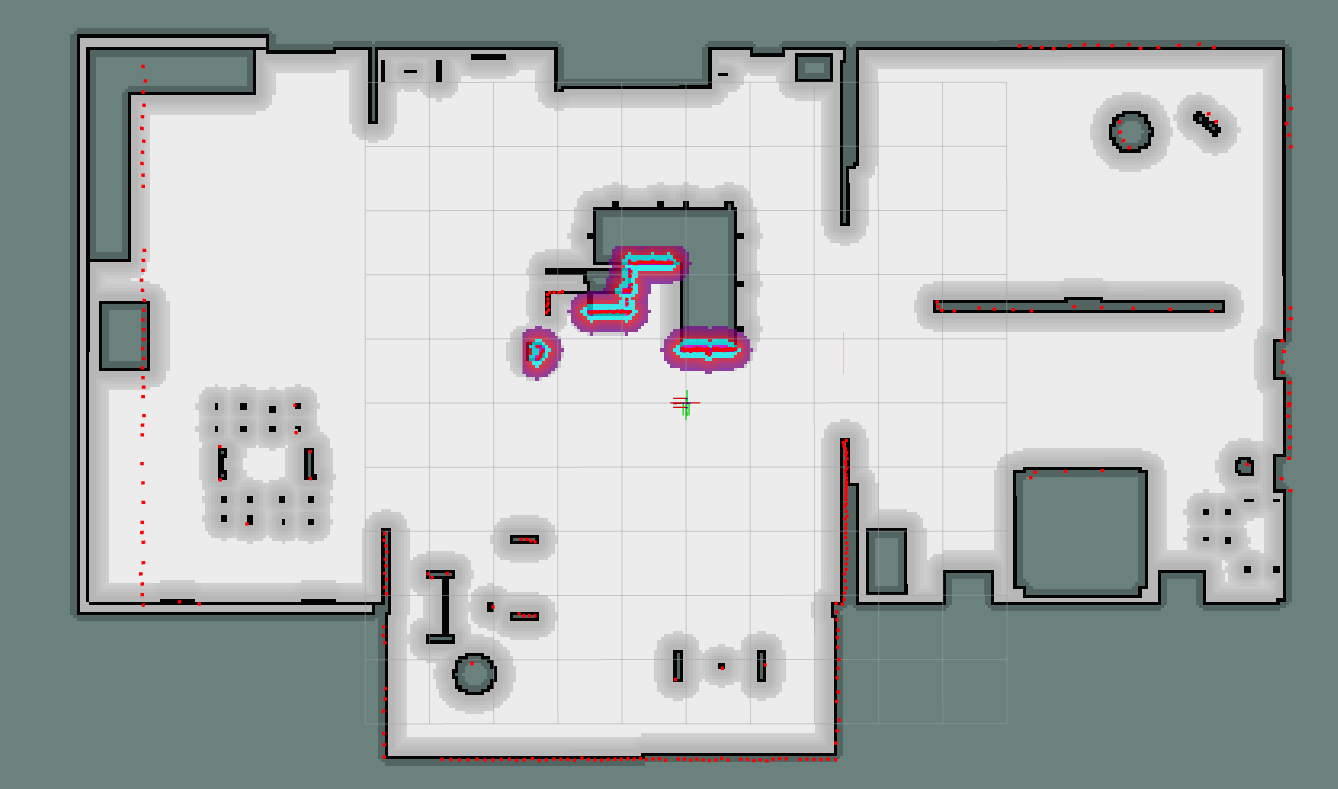}
    		\caption{RViz view}
    		\label{fig:rvi}
    	\end{subfigure}
    	\caption{Representative snapshots of the local navigation process in simulation.}
    	\label{fig:sim_demo_local_planner}
    \end{figure}
    Figure~\ref{fig:sim_demo_local_planner} shows representative Gazebo and RViz snapshots. A supplementary demonstration video is available at: \url{https://youtu.be/Sr2_hxO0ZIU}
    
    \subsubsection{Real-World Demonstration of the Local Planner}
    \label{subsec:real_demo_local}
    
    The Learning-Based DWA is further deployed on the real robot platform for real-world validation. Figure~\ref{fig:real_demo_local_planner} presents a representative snapshot of the robot navigating in the indoor environment during deployment. A supplementary demonstration video is available at: \url{https://youtu.be/3xXFdOJBP2I}
    
    \begin{figure}[htbp]
    	\centering
    	\includegraphics[width=0.6\columnwidth, trim=0 6cm 0 6cm, clip]{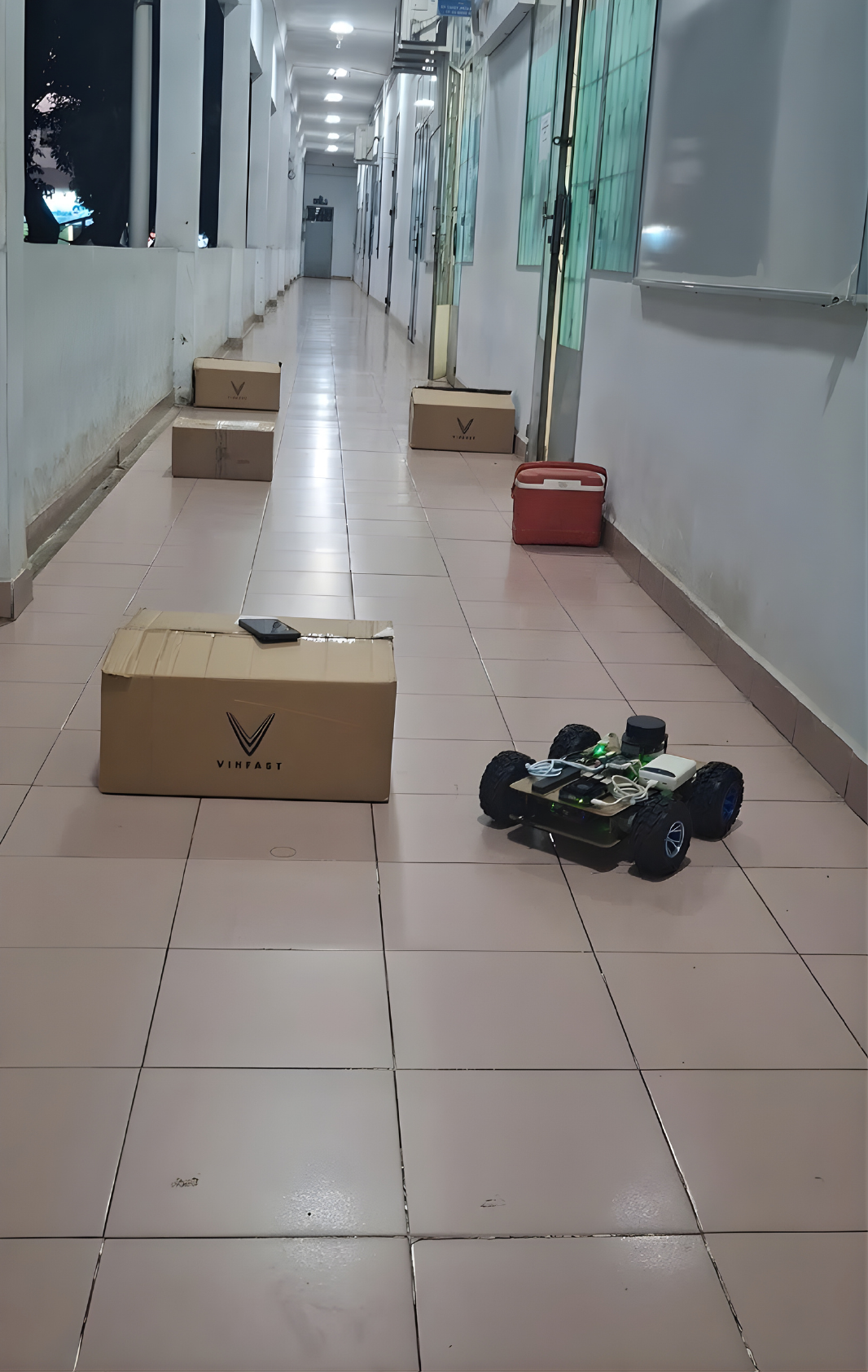}
    	\caption{Real-world deployment of the proposed local planner in an indoor environment.}
    	\label{fig:real_demo_local_planner}
    \end{figure}
    
  \section{Conclusion}
This paper presented a learning-based navigation framework for indoor mobile robots, combining a supervised neural global planner with the proposed Learning-Based DWA local planner. The global planner learns from cost-aware A* demonstrations, while the local planner selects feasible candidates from the DWA action lattice and is initialized by behavior cloning and refined with PPO under validity-aware masking. Experiments demonstrated feasible global planning, reliable local obstacle avoidance, and improved path quality and motion smoothness compared with conventional DWA, at the cost of more conservative traversal time.

Results on multiple simulated maps and one real-world indoor map provide initial evidence of generalization and feasible real-robot transfer. Further evaluation under more diverse layouts, dynamic obstacles, and sensor noise remains necessary. Future work will improve traversal efficiency and expand the training and testing environments.

\section*{Acknowledgment}
The authors acknowledge the support and facilities provided by Ho Chi Minh City University of Technology (HCMUT), VNU-HCM, for this study.
    	
    	\bibliographystyle{IEEEtran}
    	\bibliography{ref}

    \end{document}